# A Microgravity Simulation Experimental Platform For Small Space Robots In Orbit

Hang Luo, Nanlin Zhou, Haoxiang Zhang, Kai Han, Ning Zhao, Zhiyuan Yang, Jian Qi, Sikai Zhao, Jie Zhao, Yanhe Zhu*, *Member, IEEE*

*Abstract*—This study describes the development and validation of a novel microgravity experimental platform that is mainly applied to small robots such as modular self-reconfigurable robots. This platform mainly consists of an air supply system, a microporous platform and glass. By supplying air to the microporous platform to form an air film, the influence of the weight of the air foot and the ventilation hose of traditional air-float platforms on microgravity experiments is solved. The contribution of this work is to provide a platform with less external interference for microgravity simulation experiments on small robots.

## I. Introduction

In recent years, with the rapid development of space technology, space powers are vigorously developing a wide variety of space robots due to their great advantages in space exploration [1-4]. The space environment is extremely different from the ground environment. One of the most critical for machinery such as modular robots is the weightless environment. Therefore, it is very important to simulate the environment of space on the ground before modular robots work in space, to simulate and analyze the operation of modular robots in space and to conduct experiments on the ground.

The methods for constructing weightless environments on the ground are as follows: free-fall method [5]; parabolic flight method [6]; water flotation method [7,8]; air-floatation method [9-13].

With the advantages of mature technology, short construction period and unlimited experimental time, the air flotation method is a commonly used method for the construction of two-dimensional microgravity experimental platforms. Due to the characteristics of the porous material, it is often used in the construction of air feet[14-18]. For two-dimensional plane experiments, the traditional air-float method is used, the weight of the air-foot can not be ignored for small robots, and the force of the inflatable hose connected to the air-foot can not be ignored for robots, so the traditional air-foot can not simulate a good microgravity environment for small robots.

In this work, in order to allow small robots can simulate a good microgravity environment, we canceled the traditional air foot, the use of large-area porous platform, the platform above the placement of the glass, the gas supply system to the porous platform to supply gas to the porous platform to the porous platform and the glass between the resulting air film, the formation of a microgravity environment, the method and the robot is directly connected to the only one piece of glass, the quality can be ignored, and there is no longer a hose of the force of the Without the interference of the hose force, the microgravity simulation effect is greatly improved. The core contributions of this iteration are summarized as follows:

1)   Replacing traditional air feet with porous platforms and glass for good microgravity environment simulation for small robots

2)   The connection between the air supply system and the porous platform is designed to solve the problem of high pressure due to contact with high pressure air over a large area.

3)   The microgravity experimental platform is cheap and has strong universality and pertinence

4)   It was applied to the experiment of modular self-reconfiguring robot, and good experimental results were obtained.

## II. Related work

### A. Experimental Platform Principle

The construction principle of microgravity experiment platform is shown in Figure 1, which is composed of smooth glass, porous platform, gas supply system, metal platform and

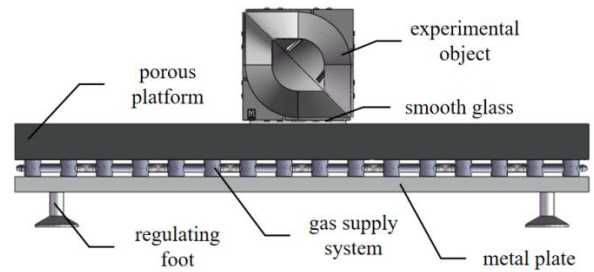

Figure 1.  Schematic Diagram of the Microgravity Experiment Platform: The platform is mainly composed of porous platform, gas supply system

*Research supported by the National Science Foundation for Outstanding Young People (accession No. 52025054) and the Key Program of the National Natural Science Foundation of China(accession No. 52435001).(Corresponding author:Yanhe Zhu)

Hang Luo, NanLin Zhou, Haoxiang Zhang, Kai Han, Ning Zhao, Zhiyuan Yang, Jian Qi, Sikai Zhao, Jie Zhao, Yanhe Zhu are with the State Key Laboratory of Robotics and System, Harbin Institute of Technology, Harbin 150001, China.(e-mail:2692530664@qq.com; zhounanlin_hit@163.com; 1259939684@qq.com; hankai9912@163.com; zhaoning1995@hit.edu.cn; yzy19962@live.com; 20b908059@stu.hit.edu.cn; zhaosikai@hit.edu.cn; jzhao@hit.edu.cn; yhzhu@hit.edu.cn).

and smooth glass.The lower part is composed of an adjusting footing and a metal plate to adjust the levelness of the porous plate.

adjusting footing. The gas supply system is connected to the lower part of the porous platform to provide high-pressure gas to the porous platform. The high-pressure gas flows out from the upper part of the porous platform. A smooth glass plate is placed on the upper part of the porous platform, and a gas film is formed between the glass plate and the porous platform. A metal plate and adjusting footing are placed under the air supply system to adjust the levelness of the porous platform.

### B. Feasibility Analysis

Due to the use of large-area porous materials, according to the traditional gas-foot construction method, if the high-pressure gas covers the lower plane of the entire porous platform, and the pressure used for half of the high-pressure gas is generally 0.4MPa, and a porous platform with 1m2 is constructed, then the force on the entire platform and the gas supply system is at least 0.4MN. With the increase of the construction area, the performance requirements of the material are higher.

In order to achieve the effect of applying porous materials to the air floating platform and solve the problem of large pressure caused by large-area porous materials, a new connection method is proposed according to the characteristics of porous materials, that is, high-pressure gas is supposed to penetrate from a small area of the porous platform platform and then exude through a larger area of the point. This can greatly reduce the area of direct contact between the high-pressure gas and the porous platform, greatly reducing the pressure. In order to verify the feasibility of the above method, we have done some theoretical analysis and experiments.

The porous material used in the porous platform is common graphite. Due to the influence of various factors on the gas flow through the porous material, such as adjusting the processing method of the porous material, the pressure provided by the air compressor, etc., what is discussed in this paper is that the gas enters the porous material from a small area and then flows out from a large area. After we pass enough gas into the center of the porous material, Points with the same distance from the center of the circle form the envelope surface. According to the isotropy of porous materials, the mass of gas flowing through each envelope surface constructed is the same. Then, according to the law of conservation of mass, the mass of gas flowing through each envelope surface in unit time is the same.We now make the following simplified analysis of the model:

1) The gas is regarded as an ideal gas, and the gas flowing between the porous material has continuity.

2) When the position difference of the gas outflow is not large, it can be simplified as the envelope surface model.

The diffusion mode of high-pressure air in porous materials is shown in Figure 2. As shown in Figure 2 (b), H is the thickness of the porous platform, and $x$ is The horizontal distance between the position of the outlet gas and the point of gas inflow. As shown in Figure 2 (a), the air intake is below the porous material and can be regarded as a point because of its small area. The air outlet is above the porous platform and flows out over the entire large area. $r_0$ is the radius of the envelope surface tangent to the surface of the porous platform, and $r$ is the radius of any envelope surface.

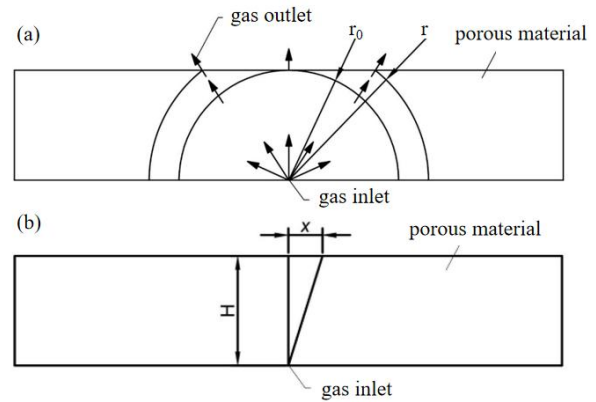

Figure 2. Example Gas Diffusion Model in Porous Materials:(a) showcases the direction of high-pressure gas flow on a porous platform and the plane of the envelope indicates intent,(b) showcases the thickness of the porous platform and a variables.

According to the above simplified model for processing, it is obvious that the temperature is almost the same everywhere, the temperature is regarded as a quantity when processing, and the mass of gas flowing through each envelope surface per unit time is the same, that is, the amount of matter is the same. Then according to the equation of state of an ideal gas, the product of pressure and flow on any envelope surface is a fixed value. When the high pressure gas flows out of the porous platform and comes into contact with the air, the pressure is the same.

Then when the radius difference of the envelope surface is not large, it is approximately regarded as the flow through each envelope surface is equal. According to the flow rate is the product of the flow rate and the area of the envelope, and the envelope surface is a sphere. Then, for the radius of the envelope surface in Figure 2 (a), the velocity on the envelope surface radius $r_0$ is assumed to be $v_0$, and the relationship between the velocity $v$, radius $r_0$, $r$ and the velocity $v_0,v$ on the envelope surface radius r is shown in

$$v \cdot r^2 = v_0 \cdot r_0^2 \tag{1}$$

According to Figure 2, the thickness of porous platform H is the same as the radius of envelope surface $r_0$, and the square of $r$ is the sum of the squares of x and H. Then the relationship between flow velocity $v_0,v$, porous platform thickness H, and horizontal distance $x$, as in

$$v = \frac{H^2}{H^2 + x^2} v_0 \tag{1}$$

According to the formula, the maximum surface flow rate on the porous platform is v0. In order to verify the accuracy of the above expression, an experiment was conducted to verify that the thickness of the porous material was set to

15mm, and after the porous platform was assembled, it was passed into high-pressure gas and placed in water. The bubbles generated were shown in Figure 3(a), proving that high-pressure gas permeated into the porous material from a small area of the porous platform. It then seeps through a larger area of points. Then, the gas flow rate was quantitatively measured for several times. Finally, the simulated diagram in the experiment was shown in Figure 3(b) with that measured in reality. It can be seen from the diagram that the experimental value is slightly smaller than the simulated value, but the trend and slope of the curve are roughly the same, so it can be seen that the approximate theoretical model constructed is reasonable.

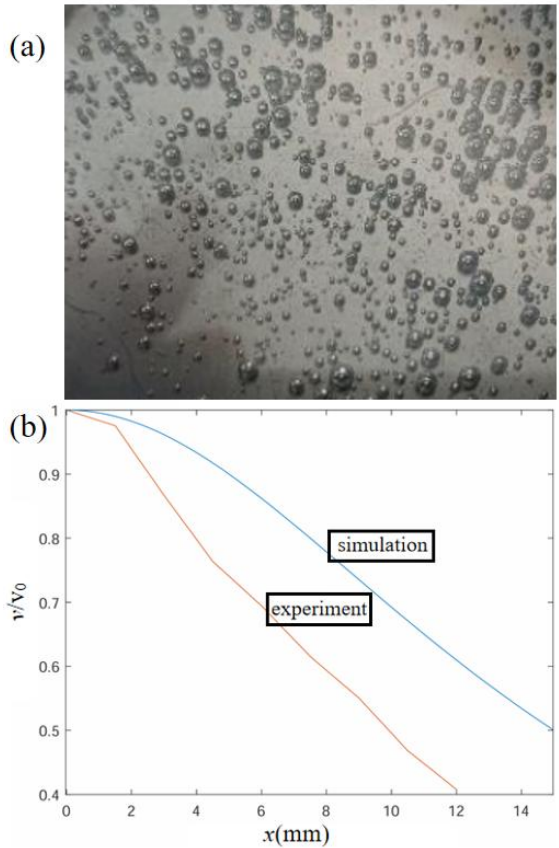

Figure 3. the Experiment of Feasibility Analysis: (a) the distribution of bubbles on the surface of the porous platform after high-pressure gas is passed into a point, (b) the relationship between the ratio of v/v0 and the horizontal distance x, the comparison between the simulated value and the experimental value is shown in the figure.

Through theoretical and physical analysis, the following conclusions can be drawn, when the envelope surface radius difference is not large, the relationship between the gas flow of a single hole can be approximated to the above relationship, if you want to get a smoother gas flow, you can appropriately increase the thickness of porous platform, increasing the thickness of porous platform can also increase the stability of the microgravity experiment platform.

*C. Construction of Microgravity Experiment Platform*

First of all, for the construction of porous platforms, graphite is a common porous material, with fine micropores, often used in the construction of air flotation platform, our choice of porous platform materials for graphite, the porosity of graphite is 17%, the particle size of 13~15μm, the most important parameters of graphite porosity and particle size, These two parameters reflect the ability to penetrate high-pressure gas, so that the ideal state of high pressure and low flow can be achieved, forming a stable gas film. Then the upper surface of the porous platform is polished, and the upper surface of the graphite is polished as smooth as possible, with high flatness and low roughness. The lower surface of the graphite is drilled, and the layout of the holes (as shown in Figure 4(a)) is arranged in a square array, and then high-pressure gas enters the porous platform from the holes.

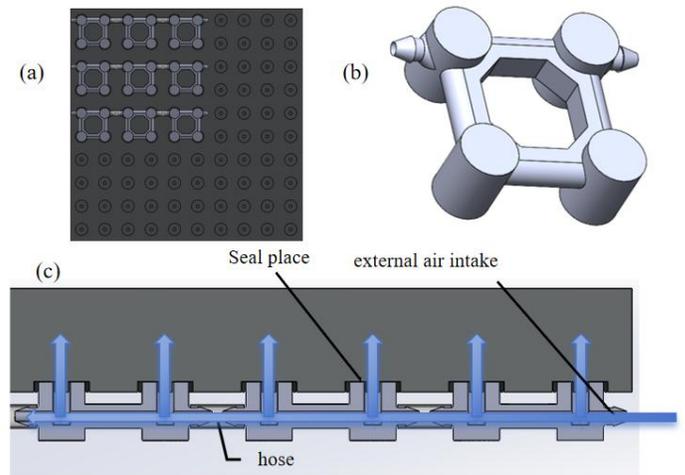

Figure 4. Porous Platform and Gas Supply System Diagram: (a) the surface intake holes under the porous platform are distributed in a square array; (b) the units of the gas supply system are obtained by 3D printing, and each unit supplies gas to four holes; hoses are used to connect the units to each other, and multiple units constitute the entire gas supply system; (c) the gas supply system is connected to the porous platform, and the arrow direction is the direction of high-pressure gas flow. The gas supply system unit and the porous platform are connected by glue.

The gas supply system is directly connected with the graphite. After many experiments and improvements, the system with good gas supply effect is finally obtained. The gas supply system is composed of a plurality of gas supply system units (as shown in Figure 4(b)), obtained by 3D printing. Each unit body supplies gas to four holes, the function is to make the unit body closely fit with the porous platform, and the units are connected by a hose. The hose is used to circulate high-pressure gas between the units, and the hose is deformable. The processing error between each unit can be eliminated, so that each unit is well connected together and closely fits the porous platform, and one end of the edge unit is directly connected with the high-pressure gas source to provide high-pressure gas for the graphite. The connection between the porous platform and the gas supply system is shown in Figure 4(c). The direction of the arrow is the direction of the high-pressure gas flow. Glue is used to connect the unit of the gas supply system and the porous platform, and the high-pressure gas enters the porous platform from a small raised cylinder. After installation, a layer of glue is poured on the lower surface of the micro-porous platform to prevent air leakage at the connection, and to prevent the gas that permeates the porous platform from flowing out from the lower surface to reduce the flow rate.

Next, a metal plate and an adjusting footing are placed under the air supply system. The area of the metal platform is the same as that of the porous platform, and its function is to adjust the levelness of the porous platform so that the experimental object is not affected by the gravity component. A piece of glass is placed on the upper surface of the porous platform, the lower surface of the glass should be low roughness and high flatness, the overall size of the glass is determined by the experimental object, and then the upper surface of the glass can be personalized design, so that it is compatible with the experimental object, without relative movement. Then, according to the overall size of the glass, the spacing of the surface under the porous platform and the size of the gas supply system unit are determined.

In the choice of graphite thickness and hole spacing under the porous platform, there are two technical routes. In order to make the microgravity experiment platform have good versatility, the porous platform can be selected as a square with a bottom side length of 2m, a graphite with a thickness of 30mm, and then the hole spacing is selected as 10mm. In this way, the microgravity experiment platform constructed can carry out the experimental needs of most small space robots. The other is for the specific space robot and specific experimental needs design, the specific process is based on the size of the space robot design with the size of the glass, and then according to the size of the glass plate design hole spacing, the general design is based on the glass plate and the porous platform contact area, in any case can be loaded with 4 arranged holes. Then design the unit body of the gas supply system according to the spacing, and finally design the size of the surface area of the porous platform according to the required experimental range, and then determine the thickness of the graphite according to the size of the surface area, the requirement is that the strength of the graphite meets the requirements without deformation, and finally determine the size of the metal plate according to the porous platform.

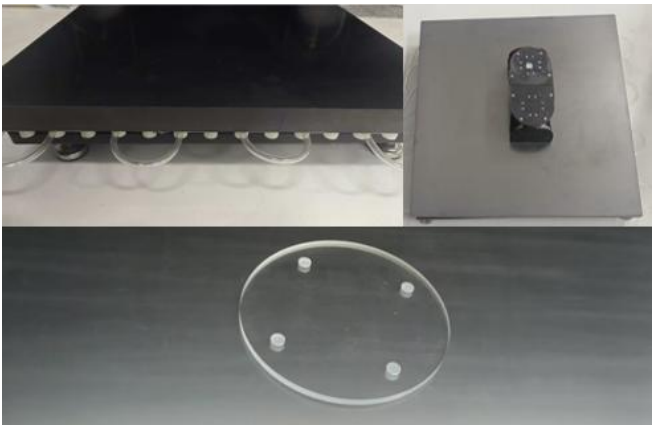

Figure 5. Microgravity Experiment Platform Physical Image: microgravity experiment platform suitable for modular self-reconfigurable robots, etc., floating glass plates and modular self-reconfigurable robots above.

According to the requirements of the laboratory, a microgravity experimental platform for modular self-reconfigurable robot Ubot was designed (as shown in Figure 5). Its basic size is 92mm×92mm×92mm. If the holes are too dense, the complexity of processing is doubled; If the holes are too sparse, the gas flow of some sides will be too low, which will have a great impact on the stability of the gas film generated. Based on the above considerations and considering the size of the modular self-reconfigurable robot, a large number of physical verification experiments are carried out for the simulation data of porous materials, and the hole spacing is finally determined to be 30mm. The thickness of the graphite used is 30mm. In the case of a modular robot with a glass of 80mm, there are generally 4 to 7 small holes of gas inflow directly in contact with the glass. After theoretical analysis and practical measurement, the microgravity experiment platform can be stable, and the simulation effect of weightlessness is very good.

III. EXPERIMENTS

The experimental phase of our study was designed to evaluate the performance and reliability of the newly developed microgravity experimental platform, as well as a number of validation experiments using modular robots. The experiment focused on three aspects:

- Microgravity simulation experiment of a single modular robot, mainly testing the mechanical rotation, rotation and linear motion of the modular robot.

- Microgravity simulation experiment of a single modular robot, mainly testing the mechanical rotation, rotation and linear motion of the modular robot.

- Microgravity simulation experiments of multiple modular robots. It mainly tests the interaction of modular robots in a weightless environment.

Through the experiments in these three aspects, the main direction of attention is to measure the performance parameters of the microgravity experimental platform, and use the microgravity experimental platform to test the modular robot in a weightless environent, and observe whether it can have good experienal results.

*A. Performance parameter test*

The microgravity experiment platform is constructed by using the above parameters. Firstly, the porous platform is levelled by the adjusting footing, so that the supporting force of the modularization is almost perpendicular to gravity. Then high-pressure gas is injected into the microgravity experiment platform, and customized glass is placed above the micro-porous platform. In this way, the microgravity platform can conduct microgravity simulation experiments for modular robots.

First of all, the measurement is the continuous running time of the microgravity experimental platform. At room temperature, the platform can continue to work stably after continuous injection of 0.4MPa gas for 1 hour, but the hose of the gas supply system connected to the micro-porous platform will be slightly hot in the later period of work, and the burning of the entire gas supply system is not obvious. Through this experiment, we can know that the microgravity experiment platform runs stably, can work continuously for a long time, and can know that the microgravity experiment platform has good stability and high safety.

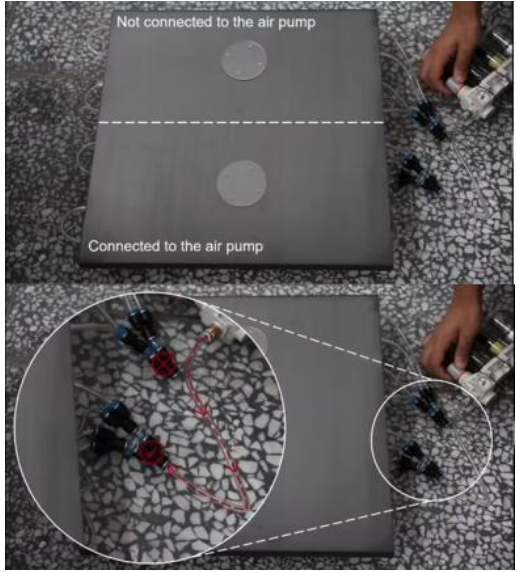

Figure 6.  Weight Loss Effect of Microgravity Experiment Platform: high pressure gas is not inserted above the porous platform, and high pressure gas is inserted below the porous platform.

Then the microgravity test platform can be measured to load the load weight. The measurement method is as follows: First, glass with a diameter of 80mm is placed on the microgravity test platform, and then weights are placed one by one above the glass. When high-pressure gas with a pressure of 0.4MPa is introduced, under the condition that a stable gas film can be formed, At this time, the carrying capacity of the microgravity experimental platform is about 100N, and the stable carrying capacity of the microgravity experimental platform per unit area is 0.02MPa after conversion.The bearing load of the microgravity experimental platform is not fixed, and is related to the material, thickness, pressure of the gas entering the porous platform, and roughness of the plane.

Under the condition that a stable gas film can be formed, the glass is directly suspended in the air and does not contact the bottom surface, so the sliding friction disappears, which can simulate a particularly good microgravity experiment effect.

Finally, a comparative experiment was carried out (as shown in Figure 6). First, a piece of glass was placed at the top and bottom respectively, and then high-pressure gas was not inserted above the porous platform, and high-pressure gas was inserted below the porous platform. After the gas is passed into the glass, a gas film is formed between the glass and the porous platform, which is very smooth when moving in the lower plane, and the movement speed can be found to be almost unchanged, and when the glass moves from the lower part of the porous platform to the top, the speed of the glass can be found to decrease significantly and finally stop. It can be seen that under the condition of forming a stable gas film, the glass is directly suspended in the air and does not contact the bottom surface, so the sliding friction disappears, which can simulate a particularly good microgravity experiment effect.

### B. Single Modular Robot Experiment

When the modular robot rotates itself (as shown in Figure 7 (a)), the porous platform on the left is fed with high-pressure gas, while the porous platform on the right is not fed with high-pressure gas. It can be seen from the experiment that when the modular robot on the left rotates, the upper part and the lower part rotate at the same Angle relative to the ground and in opposite directions, as the weight of the upper and lower parts is basically the same. That is, the whole modular robot does not rotate relative to the ground, and the direction of the modular robot before and after rotation is the same, indicating that the angular momentum of the modular robot is basically conformed. Due to the friction below the module robot on the right, the lower part of the modular robot in contact with the porous platform does not slide relative when rotating. This experiment is the same as the theoretical motion analysis of the modular robot in a weightless environment.

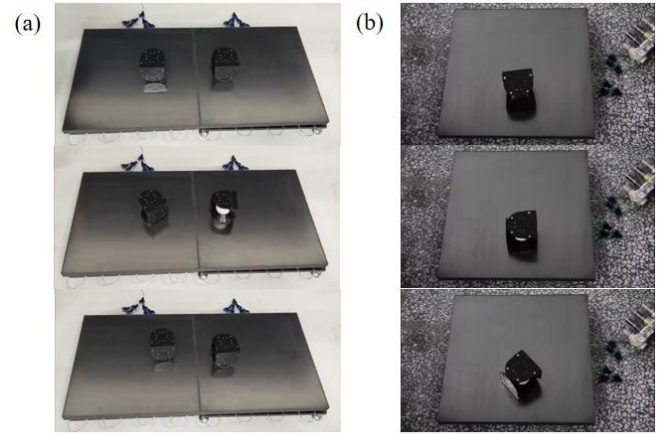

Figure 7.  a Single Modular Robot Experiment: (a) the rotation of the modular robot's own joints, the left porous platform into the high pressure gas, the right porous platform is not into the high-pressure gas, (b) the rotation of the module robot under the external force couple.

The rotation experiment of the modular robot (as shown in Figure 7 (b)) gives the modular robot a rotational torque. After the rotation of the modular robot, it can be seen that the modular robot has no tendency to stop and has been suspended in the air without sliding friction, and the angular velocity can not be seen by the naked eye.

The above two experiments show that the microgravity experimental platform can form a stable gas film for a long time, make the sliding friction disappear, and can well simulate the weightless environment in space on a two-dimensional plane.

### C. Multiple Modular Robot Experiments

The experiments of multiple modular robots are shown in Figure 8. In the initial state, the modular robots are adsorbed together due to the gravity of permanent magnets, and then the state of permanent magnet ferromagnetism is changed from gravity to repulsion. The modular robots spring off and then run at equal and opposite speeds at uniform speeds, finally reaching the edge of the porous platform at the same time.

In a weightless environment, the momentum of the system is conserved without external forces. The experimental results are the same as the theoretical analysis, indicating that the weightlessness effect of the microgravity experimental platform is good, and the platform can be used for the study of the interaction of multiple robots.

## IV. Discussion

This study presents an initial evaluation of a new microgravity experimental platform. and the experiments are carried out with modular robots. On the basis of the traditional use of air foot to build an air floating platform, our research aims to solve the limitations of using air foot to simulate the weightlessness of small in-orbit robots. Compared with the previous design, the microgravity experimental platform designed by us aims to eliminate the airfoot and smooth platform, and use porous platform and glass to replace it. Then we design an innovative connection between the gas supply system and the porous platform, and finally obtain a microgravity experimental platform with particularly strong universality and pertinence.

### A. Comparison with Prior Work

Comparing our research results with the mature air-foot platform, the traditional air-foot experiments on large satellites and space robots have negligible interference on the system due to the mass of the air-foot and the hose connected to the air-foot, which can well simulate the two-dimensional space weightlessness environment. However, for small space robots, the above-mentioned interference cannot be ignored. In view of this situation, we adopt a new way to build microgravity experiment platform. Porous platforms and glass are used instead of traditional air feet. Compared with the traditional air foot, this microgravity experimental platform has the following advantages.

The air foot and hose design is cancelled, which further reduces the interference to the microgravity experiment results. It is very friendly to small space robots. Without the interference of hoses, the weightlessness simulation can also be performed well for multiple robots working together or their interactions.

The connection design of the air supply system and the porous platform is to solve the high pressure problem caused by large area contact with high pressure air. If the traditional way is used, the high-pressure gas is directly in contact with the lower surface of the porous platform, which will lead to excessive pressure on the whole system, and the strength of the material is too high, and the design of this high-pressure gas into a small area and a large area can solve the problem of excessive pressure without affecting the weight loss effect.

The microgravity experiment platform has strong versatility and pertinence. The versatility is reflected in that the lower the spacing of the holes under the porous platform and the larger the surface area of the upper surface of the porous platform, the more kinds of robots that can be tested on the microgravity experiment platform, the stronger the versatility of the microgravity experiment platform, but the construction cost and industrial difficulty increase exponentially. Pertinence is reflected in that the size of the experimental object and the size of the experimental space can be customized to determine the appropriate parameters.

The microgravity experiment platform is cheap and simple to build. Due to the above design, the strength of the material is not high, the processing difficulty is low, so the price is cheap. Due to the design of the unit body of the gas supply system, when changing the area of the porous platform, only the number of units can be changed, and the construction of the entire microgravity experimental platform is very simple.

### B. Limitations and Future Research

Although our microgravity experimental platform has a particularly good effect on space weightlessness simulation for small space robots, it also has limitations. First of all, the upper surface of the porous platform is easy to scratch during the experiment, and it needs to be re-polished after a certain period of use. In addition, the above work is less quantitative analysis of the experimental platform, and many data are determined according to experience. Although the experimental effect is particularly good, the selection of parameters is not supported by theoretical calculation.

Future research will focus on addressing these challenges with improvements and more theoretical analysis of the microgravity experimental platform. Firstly, aiming at the problem that the surface of the porous platform is easy to scratch, a suitable film is plated on the upper surface of the direction pair to prevent scratching and make the microgravity experimental platform have higher strength. In the future, the porous platform will be simulated and modeled to obtain the relationship between gas film thickness and glass size and the parameters of hole spacing. Through theoretical analysis, the design parameters will be obtained to make the micro-gravity experimental platform have a good weight loss effect, and then the residual acceleration of the micro-gravity experimental platform will be simulated and measured. These efforts are aimed at further promoting the practical application of this new microgravity experimental platform and building a technical guide for building microgravity experimental platforms for different small space robots.

## V. Conclusion

This study developed and tested a new microgravity experimental platform for the simulation of weightless environment for small space robots, which replaced the traditional airfoot and smooth platform with porous platform and glass, and proved its ability to improve the simulation of weightless environment through experiments. Although this microgravity platform still has some limitations, this microgravity experiment platform has the advantages of universality and pertinence, and it is simple to build, low cost, and has application prospects in the simulation of weightless environment of small space robots.